\documentclass{article} 
\usepackage{nips14submit_e,times}
\usepackage{hyperref}
\usepackage{url}
\usepackage[leqno, fleqn]{amsmath}
\usepackage{amssymb}
\usepackage{qtree}
\usepackage[numbers]{natbib}
\usepackage{graphicx}
\usepackage{booktabs}
\usepackage{colortbl}
\usepackage{caption}
\usepackage{subcaption}
\usepackage{xcolor}

\usepackage[compact]{titlesec}
\titleformat{\section}{\normalsize\bfseries}{\thesection}{1em}{}
\titlespacing*{\section}{0pt}{1ex}{.5ex}

\definecolor{mylinkcolor}{rgb}{0,0,0} 
\hypersetup{colorlinks, linkcolor=mylinkcolor, urlcolor=mylinkcolor, citecolor=mylinkcolor}

\newcommand{\nateq}{\equiv}
\newcommand{\natind}{\mathbin{\#}}
\newcommand{\natneg}{\mathbin{^{\wedge}}}
\newcommand{\natfor}{\sqsubset}
\newcommand{\natrev}{\sqsupset}
\newcommand{\natalt}{\mathbin{|}}
\newcommand{\natcov}{\mathbin{\smallsmile}}

\newlength{\howlong}\newcommand{\strikeout}[1]{\settowidth{\howlong}{#1}#1\unitlength0.5ex%
\begin{picture}(0,0)\put(0,1){\line(-1,0){\howlong\divide\unitlength}}\end{picture}}

\usepackage{stmaryrd}

\usepackage{gb4e}

\def\ii#1{\textit{#1}}

\newcommand{\word}[1]{\emph{#1}}

\title{Learning Distributed Word Representations for Natural Logic Reasoning}

\author{
Samuel R.\ Bowman$^{\ast\dag}$ \\
\texttt{sbowman@stanford.edu} \\[2ex]
$^{\ast}$Stanford Linguistics \\
\And
Christopher Potts$^{\ast}$\\
\texttt{cgpotts@stanford.edu} \\[2ex]
$^{\dag}$Stanford NLP Group
\And
Christopher D.\ Manning$^{\ast\dag\ddag}$\\
\texttt{manning@stanford.edu}\\[2ex]
$^{\ddag}$Stanford Computer Science
}

\nipsfinalcopy 

\begin{document}
\maketitle

\begin{abstract}
  Natural logic offers a powerful relational conception of meaning
  that is a natural counterpart to distributed semantic
  representations, which have proven valuable in a wide range of
  sophisticated language tasks. However, it remains an open question
  whether it is possible to train distributed representations to
  support the rich, diverse logical reasoning captured by natural
  logic. We address this question using two neural network-based
  models for learning embeddings: plain neural networks and neural
  tensor networks. Our experiments evaluate the models' ability to
  learn the basic algebra of natural logic relations from simulated
  data and from the WordNet noun graph.  The overall positive results
  are promising for the future of learned distributed representations in
  the applied modeling of logical semantics.
\end{abstract}

\section{Introduction}\label{sec:intro}

Natural logic offers a powerful \emph{relational} conception of
semantics: the meanings for expressions are given, at least in part,
by their inferential connections with other expressions
\cite{vanBenthem08NATLOG,maccartney2009extended}. For instance,
\word{turtle} is analyzed, not primarily by its extension in the
world, but rather by its lexical network: it entails \word{reptile},
excludes \word{chair}, is entailed by \word{sea
  turtle}, and so forth. With generalized notions of entailment and
contradiction, these relationships can be defined for all lexical
categories as well as complex phrases, sentences, and even texts. The
resulting theories of meaning offer valuable new analytic tools for
tasks involving database inference, relation extraction, and textual
entailment.

Natural logic aligns well with distributed (e.g., vector)
representations, which also naturally model meaning relationally.
Distributed representations have been used successfully in a wide
array of sophisticated language tasks (e.g., \cite{collobert2011natural}). 
However, it remains an open question whether it is possible to train
such representations to support the rich, diverse logical reasoning
captured by natural logic; while they excel at synonymy (similarity),
the results are more mixed for entailment, contradiction, and mutual
consistency.  Using the natural logic of \cite{maccartney2009extended}
as our formal model, we address this open question using two neural
network-based models for learning embeddings: plain neural networks
and neural tensor networks (NTNs).  The natural logic is built from
the seven relations defined in Table~\ref{b-table}. Its formal
properties are now well-understood \cite{Icard:Moss:2013:LILT}, so it
provides a rigorous set of goals for our neural models. To keep the
discussion manageable, we limit attention to experiments involving the
lexicon; for a more extended treatment of complex expressions
involving logical connectives and quantifiers, see
\citet{Bowman:Potts:Manning:2014}.

In our experiments, we evaluate these models' ability to learn the
basic algebra of natural logic relations from simulated data and from
the WordNet noun graph. The simulated data help us to achieve analytic
insights into what the models learn, and the WordNet data show how they
fare with a real natural language vocabulary.  We find that only the NTN is able to fully
learn the underlying algebra, but that both models excel in the 
WordNet experiment.




\begin{table}[tp]
  \centering\small
  \setlength{\tabcolsep}{15pt}
  \renewcommand{\arraystretch}{1.1}
  \begin{tabular}{l c l l} 
    \toprule
    Name & Symbol & Set-theoretic definition & Example \\ 
    \midrule
    entailment         & $x \natfor y$   & $x \subset y$ & \ii{turtle, reptile}  \\ 
    reverse entailment & $x \natrev y$   & $x \supset y$ & \ii{reptile, turtle}  \\ 
    equivalence        & $x \nateq y$    & $x = y$       & \ii{couch, sofa} \\ 
    alternation        & $x \natalt y$   & $x \cap y = \emptyset \wedge x \cup y \neq \mathcal{D}$ & \ii{turtle, warthog} \\ 
    negation           & $x \natneg y$   & $x \cap y = \emptyset \wedge x \cup y = \mathcal{D}$    & \ii{able, unable} \\
    cover              & $x \natcov y$   & $x \cap y \neq \emptyset \wedge x \cup y = \mathcal{D}$ & \ii{animal, non-turtle} \\ 
    independence       & $x \natind y$   & (else) & \ii{turtle, pet}\\
    \bottomrule
  \end{tabular}
  \caption{The seven natural logic relations of \cite{maccartney2009extended}. 
    $\mathcal{D}$ is the universe of possible objects of the same type as those being compared, 
    and the relation $\natind$ applies whenever none of the other six do.} 
  \label{b-table}
\end{table}

\section{Neural network models for relation classification} \label{methods}

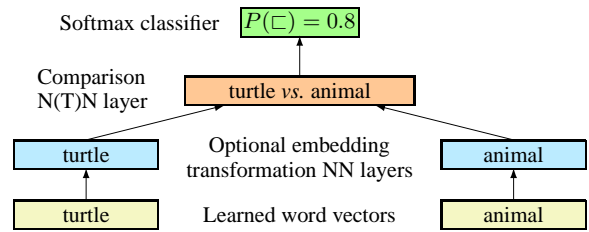
\begin{figure}[tp]
  \centering
  \footnotesize

\newcommand{\labeledtreenode}[4][3]{\put(#2){\makebox(0,0){{\fcolorbox{black}{#4}{\makebox(#1,0.3){#3}}}}}}

\newcommand{\textlabel}[4][3.5]{\put(#2){\makebox(0,0){{\makebox(#1,0.3){#3}}}}}

\definecolor{lexcolor}{HTML}{F5F7C4}
\definecolor{compositioncolor}{HTML}{BBEBFF}
\definecolor{comparisoncolor}{HTML}{FFC895}
\definecolor{softmaxcolor}{HTML}{A5FF8A}

\setlength{\unitlength}{0.61cm}

\resizebox{3in}{!}{
\begin{picture}(13.15, 5.5)
  
  \labeledtreenode[2.4]{6.5,5}{$P(\sqsubset) = 0.8$}{softmaxcolor}  
  \put(6.5,3.7){\vector(0,1){1}}  
  \labeledtreenode[5]{6.5,3.4}{turtle \emph{vs.}~animal}{comparisoncolor}

  \textlabel{2.75,5}{Softmax classifier}{black}
  \textlabel{2.0,3.4}{\parbox{2cm}{Comparison N(T)N layer}}{black}
      
  \textlabel{6.5,2.1}{Optional embedding}{black}
  \textlabel{6.5,1.5}{transformation NN layers}{black}

  \textlabel{6.5,0.5}{Learned word vectors}{black}
  

  \put(1.5,0.75){\vector(0,1){0.85}}
  \labeledtreenode{1.5,0.5}{turtle}{lexcolor}

  \put(1.5,2.25){\vector(4,1){3.25}}
  \labeledtreenode{1.5,1.9}{turtle}{compositioncolor}
  

  \put(11.5,0.75){\vector(0,1){0.85}}
  \labeledtreenode{11.5,0.5}{animal}{lexcolor}

  \put(11.5,2.25){\vector(-4,1){3.25}}
  \labeledtreenode{11.5,1.9}{animal}{compositioncolor}
  
\end{picture}
}
  \caption{The model structure used to compare \ii{turtle} and \ii{animal}.} 
  \label{sample-figure}
\end{figure}

We build embedding-based models using the method of
\cite{Bowman:Potts:Manning:2014}, which is centered on the task of
labeling a pair of words or sentences with one of a small set of
logical relations. The architecture that we use, which is limited to
only pairs of single terms (such as words), is depicted in
Figure~\ref{sample-figure}. The model represents the two input terms
as embeddings, which are fed into a comparison function based on one
of two types of neural network layer functions to produce a
representation for the relationship between the two terms. This
representation is then fed into a softmax classifier, which outputs a
distribution over possible labels. The entire network, including the
embeddings, is trained using backpropagation.

The simpler version of the comparison concatenates the two
input vectors before feeding them into a standard neural network (NN)
layer.  The more powerful neural tensor network (NTN) version uses an
additional third-order tensor parameter to allow for multiplicative
interactions between the two inputs \cite{chen2013learning}. For more
details on the implementation and training of the layer functions, see
\cite{Bowman:Potts:Manning:2014}.

This model used here differs from the one described in that work in
two ways. First, because the inputs are single terms, we do not use
a composition function here. Second, for our experiment on WordNet data, 
we introduce an additional neural network layer between the embedding 
input and the comparison function, which facilitates initializing the 
embeddings themselves from pretrained vectors and was found to help 
performance in that setting.


\section{Reasoning about semantic relations}\label{sec:join}

\begin{figure}[tp]
  \centering
  \begin{subfigure}[b]{0.42\textwidth}
    \centering  
    \setlength{\arraycolsep}{8pt}
    \renewcommand{\arraystretch}{1.1}
    \newcommand{\UNK}{\cdot}  
    \resizebox{2.2in}{!}{$\begin{array}[c]{c@{ \ }|*{7}{c}|}
        \multicolumn{1}{c}{}
        & \nateq     & \natfor     & \natrev     & \natneg    & \natalt     & \natcov     & \multicolumn{1}{c}{\natind} \\
        \cline{2-8}
        \nateq  & \nateq &   \natfor &  \natrev &  \natneg &   \natalt &  \natcov &  \natind \\
        \natfor & \natfor &  \natfor &  \UNK &  \natalt &   \natalt &  \UNK &  \UNK \\
        \natrev & \natrev &  \UNK &  \natrev &  \natcov &   \UNK &  \natcov &  \UNK \\
        \natneg & \natneg &  \natcov &  \natalt &  \nateq &    \natrev &  \natfor &  \natind \\
        \natalt & \natalt &  \UNK &  \natalt &  \natfor &   \UNK &  \natfor &  \UNK \\
        \natcov & \natcov &  \natcov &  \UNK &  \natrev &   \natrev &  \UNK &  \UNK \\
        \natind & \natind & \UNK &  \UNK &  \natind &  \UNK &  \UNK &  \UNK \\
        \cline{2-8}
      \end{array}$}
    \caption{Inference path from premises $a\,R\,b$ (row) and $b\,S\,c$ (column) to the relation that holds between $a$ and $c$, if any.  These inferences are based on basic set-theoretic truths about the meanings of the underlying relations as described in Table~\ref{b-table}. We assess our models' ability to reproduce such inferential paths.}
    \label{tab:jointable}
  \end{subfigure}
  \hfill
  \begin{subfigure}[b]{0.3\textwidth}
    \centering
    \newcommand{\labelednode}[4]{\put(#1,#2){\oval(1.5,1)}\put(#1,#2){\makebox(0,0){$\begin{array}{c}#3\\\{#4\}\end{array}$}}}
    \setlength{\unitlength}{1cm}
    \resizebox{1.5in}{!}{\begin{picture}(5,5.5)
      \labelednode{2.50}{5}{}{1,2,3}
      
      \put(0.75,4){\line(3,1){1.5}}
      \put(2.5,4){\line(0,1){0.5}}
      \put(4.25,4){\line(-3,1){1.5}}
      
      \labelednode{0.75}{3.5}{a,b}{1,2}
      \labelednode{2.50}{3.5}{c}{1,3}
      \labelednode{4.25}{3.5}{d}{2,3}
      
      \put(0.75,2.5){\line(0,1){0.5}}
      \put(0.75,2.5){\line(3,1){1.5}}
      
      \put(2.5,2.5){\line(-3,1){1.5}}
      \put(2.5,2.5){\line(3,1){1.5}}
      
      \put(4.25,2.5){\line(0,1){0.5}}
      \put(4.25,2.5){\line(-3,1){1.5}}

      \labelednode{0.75}{2}{e,f}{1}
      \labelednode{2.50}{2}{}{2}
      \labelednode{4.25}{2}{g,h}{3}
      
      \put(2.5,1){\line(-3,1){1.5}}
      \put(2.5,1){\line(0,1){0.5}}
      \put(2.5,1){\line(3,1){1.5}}
      
      \labelednode{2.5}{0.5}{}{}
    \end{picture}}
    \caption{Simple boolean structure. The letters name the sets. Not all sets have names, and
    some sets have multiple names, so that learning $\nateq$ is non-trivial.}\label{lattice-figure}
  \end{subfigure}
  \hfill
  \begin{subfigure}[b]{0.2\textwidth}
    \centering
    \setlength{\tabcolsep}{12pt}
    \resizebox{1.05in}{!}{\begin{tabular}[c]{c  c}
      \toprule
      Train & Test \\
      \midrule
                    & $b \nateq b$ \\
      $b \natcov c$ &               \\
                    & $b \natcov d$ \\
                    & \strikeout{$b \natrev e$} \\
      $c \natcov d$ &               \\
      $c \natrev e$ &               \\
                    & \strikeout{$c \nateq f$} \\
      $c \natrev g$ &               \\ 
                    & $e \natfor b$ \\
      $e \natfor c$ &               \\[-1ex]
      $\vdots$      & $\vdots$ \\
      \bottomrule
    \end{tabular}}

    \caption{A train/test split of the atomic statements about the
      model.  Test statements not provable from the training data are
      crossed out.}\label{unprovable}
  \end{subfigure}  
  \caption{Experimental goal and set-up for reasoning about semantic relations.}
  \label{exp1}
\end{figure}
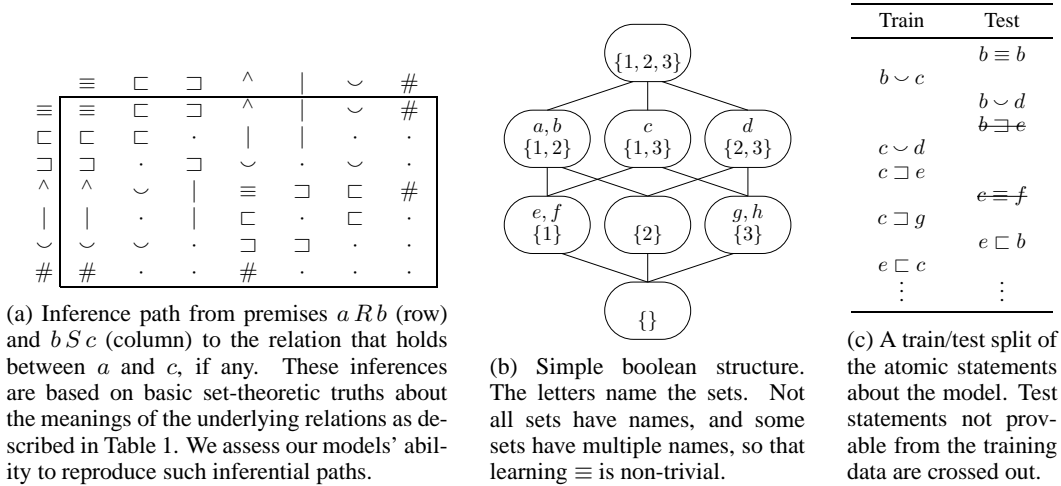

In this experiment, we test our model's ability to learn and use
the natural logic inference rules schematized in 
Figure~\ref{tab:jointable}. For instance, given that $a \natrev b$ and $b
\natrev c$, one can conclude that $a \natrev c$, by basic
set-theoretic reasoning (transitivity of $\natrev$). Similarly, from
$a \natfor b$ and $b \natneg c$, it follows that $a \natalt c$. Cells in the
table containing a dot correspond to pairs of relations for which no valid 
inference can be drawn in our logic.


\paragraph{Experiments}
To test our models' ability to learn these inferential patterns, we
create artificial boolean structures in which terms denote sets of entities from
a small domain (e.g., Figure~\ref{lattice-figure}), employ logical
deduction to identify the valid statements, divide those into train
and test sets, and remove from the test set those statements which
cannot be proven from the training statements
(Figure~\ref{unprovable}).
%
%
%
In our experiments, we create 80 randomly generated sets drawn from a
domain of seven elements. This yields 6400 statements about pairs of
formulae, of which 3200 are chosen as a test set, and that test set is
further reduced to the 2960 examples that can be provably derived from
the training data. We trained both the NN model and the NTN model on
these data
sets.

\paragraph{Results} 
We found that the NTN model worked best with 11-dimensional vector
representations for the 80 sets and a 90-dimensional feature vector
for the classifier, though the performance of the model was not highly
dependant on either dimensionality setting. 
Averaging over five randomly generated data sets, the model was able to correctly label 98.1\% (standard error $0.67\%$) of the provably derivable test examples, and 87.7\%
($\textit{SE} = 3.59\%$) of the remaining test examples. The simpler NN worked
best with 11 and 75 dimensions, respectively, but was able to achieve
accuracies of only 84.8\% ($\textit{SE} = 0.84\%$) and 68.7\% ($\textit{SE} = 1.58\%$),
respectively. Training accuracy was 99.8\% ($\textit{SE} = 0.04\%$) for the NTN and 94.7\% ($\textit{SE} = 0.89\%$) for
the NN.

These results are fairly straightforward to interpret. The NTN model
was able to accurately encode the relations between the terms in the
geometric relations between their vectors, and was able to then use
that information to recover relations that were not overtly included
in the training data. In contrast, the NN was able to achieve this
behavior only incompletely. It is possible but not likely that it
could be made to find a good solution with further optimization on
different learning algorithms, or that it would do better on a larger
universe of sets, for which there would be a larger set of training
data to learn from, but the NTN is readily able to achieve these
effects in the setting discussed here.

\section{Reasoning about lexical relations in WordNet}\label{sec:wordnet}

Using simulated data as above is reassuring about what the models
learn and why, but we also want to know how they perform with a real
natural language vocabulary. Unfortunately, as far as we are aware,
there are no available resources labeling such a vocabulary with the
relations from Table~\ref{b-table}. However, the relations in WordNet
\cite{fellbaum2010wordnet} come close and pose the same substantive
challenges within a somewhat easier classification problem.


We extract three types of relation from WordNet. \ii{Hypernym} and
\ii{hyponym} can be represented directly in the WordNet graph
structure, and correspond closely to the $\natrev$ and $\natfor$
relations from natural logic. As in natural logic, these relations are
mirror images of one another: if \ii{dog} is a hyponym of \ii{animal}
(perhaps indirectly by way of \ii{canid}, \ii{mammal}, etc.), then
\ii{animal} is a hypernym of \ii{dog}. We also extract \ii{coordinate}
terms, which share a direct hypernym, like \ii{dalmatian}, \ii{pug},
and \ii{puppy}, which are all direct hyponyms of \ii{dog}.  Coordinate
terms tend to exclude one another, thereby providing a loose approximation 
of the natural logic exclusion relation $\natalt$. 
WordNet defines hypernymy and hyponymy over
sets of synonyms, rather than over individual terms, so we do not
include a \ii{synonym} or \ii{equivalent} relation, but rather
consider only one member of each set of synonyms. Word pairs which do
not fall into these three relations are not included in the data set.

To limit the size of the vocabulary without otherwise simplifying the learning problem, we extract all of the
instances of these three relations for single word nouns in WordNet that are hyponyms of the node 
\texttt{organism.n.01}. In order to balance the distribution of the classes, we slightly downsample instances 
of the \ii{coordinate} relation, yielding a total of 36,772 relations among 3,217 terms. We report results below using crossvalidation, choosing a disjoint 10\% test sample for each of five runs. Unlike in the previous experiment,
it is not straightforward here to determine in advance
how much data should be required to train an accurate model, so we performed training runs with 
various fractions of the remaining data. Embeddings were fixed at 25 dimensions and were initialized 
randomly or using distributional vectors from GloVe \cite{pennington2014glove}. The feature vector 
produced by the comparison layer was fixed at 80 dimensions.

\paragraph{Results} 
We find that the NTN performs perfectly with random initialization, and that the plain NN performs almost as well,
a point of contrast with the results of Section~\ref{sec:join}. We also find that initialization with GloVe is helpful in allowing the models to maintain fair performance
with smaller amounts of training data. Some of the randomly initialized model runs failed to learn
usable representations at all and labeled all examples with the most frequent labels. We excluded these runs from the statistics, but marked settings for which this occurred with the symbol $\dagger$. For all of the remaining runs, training accuracy was consistently above 99\%.

\newcommand{\nodagger}{\phantom{$^\dagger$}}

\begin{table}[h]
\centering\resizebox{5.5in}{!}{
  \setlength{\tabcolsep}{15pt}
  \renewcommand{\arraystretch}{1.1}
  \begin{tabular}{r r@{ }r r@{ }r r@{ }r r@{ }r r@{ }r} 
    \toprule
     Portion of & \multicolumn{4}{c}{NN} & \multicolumn{4}{c}{NTN} & \multicolumn{2}{c}{Baseline}\\
     training data  & \multicolumn{2}{c}{w/ GloVe} & \multicolumn{2}{c}{w/o GloVe} & \multicolumn{2}{c}{w/ GloVe} & \multicolumn{2}{c}{w/o GloVe} \\
    \midrule
     100\%   & 99.73 &(0.04) & 99.37$^\dagger$ &(0.14)    & 99.61 &(0.02) & \textbf{99.95}\nodagger &(0.03) & 37.05 &(--)\\
     33\%    & 95.96 &(0.20) & 95.30\nodagger &(0.12)                & 95.83 &(0.35)          & 95.45$^\dagger$ &(0.31) & 37.05 &(--)\\
     11\%    & 91.11 &(0.24) & 90.81$^\dagger$ &(0.20)    & 91.27 &(0.27)          & 90.90$^\dagger$ &(0.13) & 37.05 &(--)\\
    \bottomrule
  \end{tabular}}
 \caption{Mean test \% accuracy scores (with standard error) on the WordNet data over five-fold crossvalidation. The baseline figure is the frequency of the most frequent class, \ii{hypernym}.\label{wn-table}} 
\end{table}

\section{Conclusion}\label{sec:discussion}

This paper evaluated two neural models on the task of learning 
natural logic relations between distributed word representations. The
results suggest that at least the neural tensor network
has the capacity to
meet this challenge with reasonably-sized training sets, learning both 
to embed a vocabulary in a way that encodes a diverse
set of relations, and to subsequently use those embeddings to infer new
relations. In
\cite{Bowman:Potts:Manning:2014}, we extend these results to include
complex expressions involving logical connectives and quantifiers, with similar
conclusions about (recursive versions of) these models. These findings
are promising for the future of learned distributed representations in the
applied modeling of logical semantics.

\bibliographystyle{unsrtnat}

\bibliography{MLSemantics} 

\end{document}